\documentclass[
twocolumn,
]{ceurart}

\usepackage{listings}

\usepackage{amsmath}

\usepackage{array}
\usepackage{tabularx}

\usepackage{times}
\usepackage{mathptmx}
\usepackage{latexsym}
\usepackage{hyperref}
\usepackage[noabbrev,capitalize,nameinlink]{cleveref}
\usepackage{xcolor, soul, colortbl}
\usepackage{array, multirow, graphicx}
\usepackage{adjustbox}
\usepackage[T1]{fontenc}
\usepackage{booktabs}
\usepackage{tikz}
\usepackage{tikz-dependency}
\usepackage{lscape, float} 
\usepackage{changes}
\usepackage[shortlabels]{enumitem}

\AtBeginEnvironment{table*}{\fontfamily{ptm}\selectfont}
\AtEndEnvironment{table*}{\fontfamily{ptm}\selectfont}

\usepackage{newfloat}
\DeclareFloatingEnvironment[fileext=lop]{tablecustom}
\floatname{tablecustom}{Table}

\newcommand*\circled[1]{\tikz[baseline=(char.base)]{
            \node[shape=circle,draw,inner sep=.6pt] (char) {#1};}}

\begin{document}

\copyrightyear{2023}
\copyrightclause{Copyright for this paper by its authors.
  Use permitted under Creative Commons License Attribution 4.0
  International (CC BY 4.0).}

\conference{ArXiV Pre-Print}

\title{Large Language Models as Batteries-Included Zero-Shot ESCO Skills Matchers}

% \tnotetext[1]{You can use this document as the template for preparing your
%   publication. We recommend using the latest version of the ceurart style.}

\author[1]{Benjamin Clavi\'{e}}[%
orcid=0000-0002-8346-8703,
email=ben.clavie@brightnetwork.co.uk,
]
\cormark[1]
\author[2]{Guillaume Souli\'{e}}[%
email=guillaume.soulie@brightnetwork.co.uk,
]
\address[1]{Bright Network Ltd., Edinburgh, UK}

% \author[1]{Anonymised Authors}[%
% ]
% \address[1]{Anon}
% \cormark[1]

% \author[4]{Manfred Jeusfeld}[%
% orcid=0000-0002-9421-8566,
% email=Manfred.Jeusfeld@acm.org,
% url=http://conceptbase.sourceforge.net/mjf/,
% \address[4]{University of Skövde, Högskolevägen 1, 541 28 Skövde, Sweden}

\cortext[1]{Corresponding author.}

\begin{abstract}
Understanding labour market dynamics requires accurately identifying the skills required for and possessed by the workforce. Automation techniques are increasingly being developed to support this effort. However, automatically extracting skills from job postings is challenging due to the vast number of existing skills. The ESCO (European Skills, Competences, Qualifications and Occupations) framework provides a useful reference, listing over 13,000 individual skills. However, skills extraction remains difficult and accurately matching job posts to the ESCO taxonomy is an open problem.
In this work, we propose an end-to-end zero-shot system for skills extraction from job descriptions based on large language models (LLMs). We generate synthetic training data for the entirety of ESCO skills and train a classifier to extract skill mentions from job posts. We also employ a similarity retriever to generate skill candidates which are then re-ranked using a second LLM. Using synthetic data achieves an RP@10 score 10 points higher than previous distant supervision approaches. Adding GPT-4 re-ranking improves RP@10 by over 22 points over previous methods. We also show that Framing the task as mock programming when prompting the LLM can lead to better performance than natural language prompts, especially with weaker LLMs.
We demonstrate the potential of integrating large language models at both ends of skills matching pipelines. Our approach requires no human annotations and achieve extremely promising results on skills extraction against ESCO.
\end{abstract}


\maketitle

\section{Introduction}

The Job Market is often described as constantly evolving, as technological developments and societal changes result in large changes in its makeup, as has been frequently studied and demonstrated \cite{jobmarket1, jobmarket2}. In recent years, the rapid digitisation of society has led to entirely new categories of skills becoming requirements for many jobs\cite{digitalisation}. As demands for skills evolve, there is an increasing need to better understand the skills required by jobs. This has supported the continuous development of skill taxonomies such as the European Union's European Skills, Competences, Qualifications and Occupations, or \textbf{ESCO} \cite{ESCO}, framework, developed to improve understanding and efficiency of the wider EU job market.

While useful, such frameworks require \textbf{skill extraction} (SE) approaches to understand skills present in job postings and enable automation at scale. Skills Extraction has recently been the subject of an increased amount of interest\cite{escoxlmr}, which becomes even more important as research shows that a large proportion of required skills are implicitly expressed rather than explicitly stated within postings \cite{bhola}.

However, automated skills extraction faces considerable roadblocks. Most notably, many efforts are limited in scale by the lack of available data for both training and evaluation. Recent efforts have begun to try and alleviate this effort, by publicly releasing manually annotated datasets. These approaches, while promising, suffer from the complexity of the task, as the ESCO taxonomy contains 13 890 individual skills. As a result, the task is often re-framed or simplified, for example by converting the task into a span-extraction task \cite{skillspan, sayfullina}, leaving direct matching to the taxonomy for future work. Some recent work has explored this Extreme Multi-Label Classification (XLMC) task with an end-to-end approach to matching and extraction with non-ESCO taxonomies, with encouraging results \cite{bhola, jobxlmc}, although sometimes relying on simplifying the taxonomy by using only higher-level labels \cite{cnnskills}.

In fact, the task of \textit{Skills Extraction \textbf{against a taxonomy}} could be framed as a two-tasks process: \circled{1} an \textbf{extraction} step, focused on recognising the potential mentions of skills, or groups of skills, from the content of job postings, on which strong progress has been made \cite{skillspan, fijo, bhola}, and \circled{2} a \textbf{matching} step, akin to extreme multi-label classification, focused on linking these mentions with fine-grained taxonomies, which remains a difficult problem.

The sheer number of existing skills makes it very difficult to obtain sufficient training data for comprehensive coverage. As such, various techniques, such as using the ESCO API as a form of distant supervision \cite{kompetencer} and generating training data through such distant supervision techniques \cite{negsample}, have been explored.

Meanwhile, the rapid development and improvement of generative large language models (LLMs) \cite{llmsurvey} and especially instruction-tuned LLMs \cite{flan, instructgpt} , models further trained to specifically follow natural language instructions,  has resulted in the use of generative LLMs on a large array of applications, often yielding competitive or even state-of-the-art results across many tasks, as highlighted by the release and strong performance of OpenAI's GPT-4 \cite{gpt4}. Notably, LLMs have shown their ability to improve performance on text retrieval tasks through the generation of synthetic training data based on a handful of real examples \cite{inpars}. They have also shown considerable problem-solving ability, often anthropomorphised as \textit{reasoning}, which appears to be even stronger when the task is framed as a programming task \cite{programmingreasoning}.

Due to the nature of the training data used by Large Language Models, which contains very large volumes of content scraped from across the internet \cite{commoncrawl}, we hypothesise that the knowledge representation embedded within the models makes them particularly suitable for broader job understanding tasks, as many job postings, as well as skills descriptions, were present in this data. As such, we will explore the creation of a \textbf{zero-shot skills matching} pipeline through the use of large language models, focusing in this case on the use of GPT-3.5 and GPT-4.

\paragraph{}{\textbf{Contributions}} In this paper, we show that:
\circled{1} Large Language Models can reliably generate zero-shot training data that improves performance in the skill matching task. We then show that using this data to power both similarity-based retrieval approaches and linear classifier models trained on the data can generate good lists of potential skills found within given text extracts, outperforming the previous approaches.
\circled{2} The \textbf{Skill Matching} task can be framed as a two-step problem, with the first stage consisting of generating a list of potential skill matches and the second stage focusing on re-ranking these potential matches. We show that LLMs can be used as zero-shot rerankers for this second step of the extraction pipeline with very strong performance without the need for annotated data.
\circled{3} Framing the skills matching task as a mock programming problem provides a further performance boost for both large language models tested. More notably, it improves the performance of the less capable model by over 10 percentage points by enhancing its ability to follow instructions.

% \section{Background}

% \paragaph{}{\textbf{Large Language Models}} hello test line.

% \paragaph{}{\textbf{Skill Matching}} hello test line.

\section{LLM-Generated Data}
As discussed, we follow a two-step process to form the \textbf{Skills Matching pipeline}. While, as we will demonstrate later, LLMs are strong zero-shot rerankers, it's impossible for them to work as standalone classifiers for all ESCO skills, as the 13890 skills listed in ESCO do not fit within the context window of most Language Models. Even if this were possible, the increased compute and processing time required by current Transformer-based models (used by all LLMs) would make this approach unsuitable in many cases, despite promising advancements in more computationally efficient approaches to Transformer models \cite{fastattention}.

We thus choose to leverage large generative models for the skills matching task through the generation of \textbf{synthetic training data}. We believe that this synthetic training could allow us to perform the skills matching task without having to reframe the task or rely on useful but limited distant supervision techniques\cite{negsample}.

For each of the 13890 skills contained in ESCO, we prompt \textsc{GPT-3.5} \footnote{Our limited experiments showed that all recent LLMs, including the open-source Flan-UL2 \cite{flanul2}, could potentially perform this task without a strong negative impact on performance. Thoroughly evaluating different LLMs for this data generation step is beyond the scope of this work, but would likely be a valuable future area of research, as we noticed considerable style differences between different models, leading us to believe a combination of different ones could generate a more diverse training data set.} to generate forty example sentences that could be used in a job posting in order to refer to the skill. We specifically request that the sentences be phrased in a variety of ways, and be of various lengths (from just a few words directly referring to the skill to a few sentences mentioning it implicitly). We provide slightly different instructions based on the "skill type", for instance requesting explicit mentions in more examples when generating data for a skill contained in the \textbf{tech} ESCO skill listing, as programming languages tend to be clearly mentioned in job ads. We also use additional information or skill descriptions present in the ESCO data to enrich our prompt and help the model disambiguate between potentially ambiguous terms.

Once the training data is generated, it is not thoroughly manually reviewed. We performed programmatic checks, showing that a full forty examples were generated for more than 97\% of skills, with a few of them having fewer examples due to model context size limits or failure to properly follow instructions, which were not addressed, as the entirety of skills but one \footnote{The skill "semen insertion", related to veterinary work, could not have training data generated for it by the GPT model family, as its name triggers OpenAI's content filters.} had more than 30 generated examples. We sampled a random 100 skills for manual review to ensure that the generated data met our criteria and that the prompts contained no obvious mistakes.

The prompts used for data generation are provided in appendix \ref{app:A}.

\section{Potential Skills Identification}
Following data generation, our next step is the identification of potential skills contained within a given text span. We generate these potential skills through two main approaches: a \textbf{linear classifier-based approach}, using linear regression classifiers on frozen embeddings, and \textbf{textual similarity approaches}, where we rely on the cosine distance between embeddings to determine whether a skill is potentially present or not.

For both of these approaches, we use the \textsc{e5-large-v2} text embedding model \cite{e5}, the current state-of-the-art embedder for similarity-based information retrieval whose generated embeddings have also been shown to reach strong performance for few-shot text classifications, making it particularly suitable for both of our approaches.

\subsection{Classifier Candidates}

Our first set of candidates is generated by the use of simple logistic regression classifiers. We train one binary classifier per label, in a one-versus-all classification approach. We use no real-world data for these classifiers, instead training only on the synthetic data generated as described in section 3.For any given class, we treat all example sentences generated for the label as positive examples and sample twice as many examples from other labels to use as negative examples.

Following previous work on skill classification \cite{negsample}, we use partial hard negative sampling \cite{hardneg} to ensure our models are better at distinguishing between very similar labels. To do so, we make it so 10\% of the negative examples are hard negatives, sampled from labels associated with the example sentences having the highest cosine similarity with the positive examples.

At inference time, we consider every positive classification from the classifiers as a candidate label to be used for re-ranking.

\subsection{Similarity-based Candidates}

Our second approach to generating candidate is based on \textbf{cosine similarity} between embeddings. We generate candidate through two distinct approaches: \textbf{label similarity} and \textbf{sentence similarity}.

\paragraph{}{\textbf{Label Similarity}} This is a simple similarity look-up between a target extract and the full list of existing labels. We do not set a threshold for this step, rather, we treat the 40 most similar labels as candidates to be provided to the re-ranker.
\paragraph{}{\textbf{Sentence Similarity}} For this candidate generator, we use the cosine distance between the current extract and the synthetic example sentences. If two sentences with the same label are part of the 40 most similar sentences, we add the label to the list of candidates. This is in effect similar to a simplified form of k-nearest neighbour classification \cite{knn}.

\section{Zero-Shot Potential Skills Ranking}

Once the list of potential skills found in a given span is generated, we prompt an LLM to extract and rank the ten most likely skills in order of suitability. This is the \textbf{reranking} step, a key component of information retrieval pipelines, which is increasingly performed by fine-tuned language models \cite{rerank}. In this work, we explore how a zero-shot approach leveraging LLMs' "learned knowledge" performs on our task.

\subsection{A Note About the Evaluated LLMs}
We report results for both \textsc{GPT 3.5}, an instruction-tuned \cite{instructgpt}, more powerful version of GPT-3 \cite{gpt3}, more specifically\textsc{gpt-3.5-turbo-0301} and \textsc{GPT 4} \cite{gpt4} (\textsc{gpt-4-0314}), which produced the most promising results in our exploratory work. Most of our prompt engineering work was performed for \textsc{GPT 4}, and re-used as-is for \textsc{GPT 3.5}.

While we have not conducted extensive experiments using them, our exploratory work has shown that open-source LLMs, of which \textsc{Flan-UL2} \cite{flanul2} was the best performing at the time of this work, failed to produce reliable outputs, frequently "hallucinating" skills, in a way similar to \textsc{GPT 3.5}. However, in the case of \textsc{GPT3 3.5}, this was entirely mitigated by the \textit{Python approached} described below.

At the time of conducting our experiments, \textsc{Falcon} \cite{falcon} had not yet been released and the authors did not have access to \textsc{LLaMa} \cite{llama}.

\subsection{LLMs as Reranker}

We use prompting and prompt engineering \cite{prompteng}and describe the task or reranking in the prompt. We use a chat-formatted prompt, through OpenAI's ChatML \cite{chatml}. We give the model a broad description of its role as its initial prompt, followed by the detailed instructions for the task, and a mocked message from the model acknowledging and summarising the instructions.

We then provide the model with a list of potential skills, generated by the previously described methods. We experiment with both providing the model with information about the score it received as a potential skill, through either classification class probability or textual similarity, depending on the potential skill's source. We found this had no impact on performance, and therefore do not provide this information to the model in our final evaluation to reduce the number of token used in our prompts, thus reducing the required compute. We pass all skills identified by the \textsc{classifier} approach as well as up to 60 skills identified by the \textsc{similarity-based approach}.

We request an ordered list of the ten most likely skill matches in our prompt. In all cases, we provide the model with the ability to use the \textit{NO\_LABEL} skill to reach 10 skills it identifies fewer or no matches.

All the prompts used for the reranking task are provided in Appendix \ref{app:A}.

\subsubsection{Mock Python Programming Variant}
Recent work has shown that Large Language Models can often perform better on "reasoning" tasks when they are approached as programming exercises \cite{programmingreasoning}. Additionally, anecdotal evidence often states that it is easier to control the output of large language models when requiring programming-like outputs, supposedly due to programming languages' more structured nature. While investing the full extent of these claims is beyond the scope of this work, we experiment with modifying our re-ranking prompt to include explicit instructions to answer exclusively in Python, in the form of a function returning an ordered list of ranked skills, and outputting the justification for inclusion as a comment. No other modifications to the instructions are made.

We choose Python over other programming languages as it is often a good proportion of the programming data commonly used to train and evaluate LLMs \cite{codex, palm} and requires very little adjustments to the existing re-ranking step presented in the previous section, which is itself implemented in Python.

\section{Experimental Setup}

\begin{figure*}[ht]
  \centering
  \includegraphics[width=\textwidth]{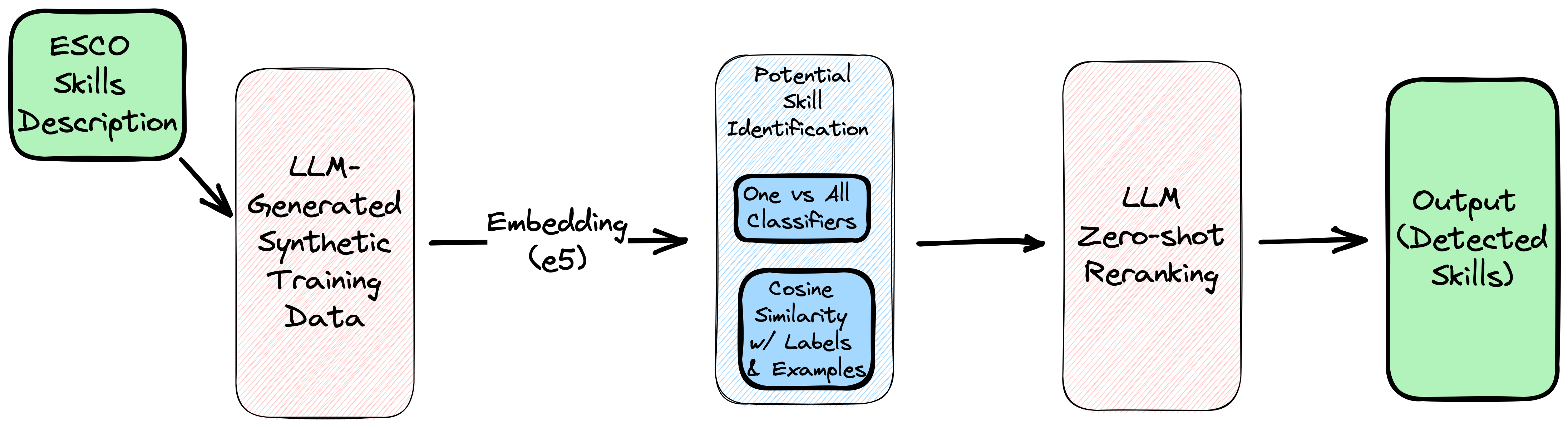}
  \caption{High-level overview of the full process.}
  \label{arch}
\end{figure*}

As ESCO-based skills matching is an Extreme Multi-Label Classification (XMLC) task, we choose to frame it similarly to an information retrieval task and use IR-inspired methods, for which we provide a high-level overview of our architecture in Figure \ref{arch}. Skills matching against the ESCO taxonomy, due to the very large number and granularity of skill labels, justifies this framing: while we want to assign as many labels present in our test set as possible, it is also highly likely that many potentially relevant labels are not attached to our test examples, either because of oversight or because of subjective judgement in situations where multiple similar labels applied. As such, our aim is to maximise our retrieval of test labels, without harsh penalties for additional labels assigned by the model.

\subsection{Evaluation}

\subsubsection{Data} We evaluate our approach on the dataset provided by Decorte et al. \cite{negsample}. Their work built upon the SkillSpan dataset\cite{skillspan}, a publicly available dataset focused on the detection of text spans containing the mention of either \textit{skills} or \textit{knowledge}, which are two sub-categories of \textit{skills} as broadly defined within the ESCO framework. Using the extracted spans, Decorte et al. manually assigned ESCO skills to the extracted spans in order to create a dataset of spans annotated with the matching ESCO skill(s). To the best of our knowledge, this represents the best effort at creating an evaluation dataset using the full extent of ESCO's fine-grained approach rather than approximations or groupings. We use the \textit{validation} set provided to tweak our prompts and evaluate our models on the \textit{test} set.

The data contains two distinct subsets, \textit{TECH}, which focuses on data extracted from jobs within the tech sector, and \textit{HOUSE}, containing more generalist jobs. We report results for each set separately, following the literature.

\subsubsection{Metrics} As per the work introducing our evaluation dataset \cite{negsample}, we report the macro-averaged \textbf{R-Precision@k (RP@k)}, which is particularly well-suited to evaluated extreme multilabel classification tasks such as this one \cite{chalkidis, negsample} as well as the \textbf{Mean Reciprocal Rank (MRR)} of the highest ranked correct label as a further indication of ranking quality.

% Additionally, we report the Hit Rate@k (HR@k), which measures the percentage of instances in which at least one of the \textit{k} top ranked labels was correct.

\section{Results and Discussions}

\begin{tablecustom*}[]
\begin{tabular}{l|cccccccc}
                               & \multicolumn{4}{c}{\textbf{House}}                                                                                           & \multicolumn{4}{c}{\textbf{Tech}}                                                                         \\
                               & \multicolumn{1}{l}{MRR} & \multicolumn{1}{l}{RP@1} & \multicolumn{1}{l}{RP@5} & \multicolumn{1}{l|}{RP@10}                   & \multicolumn{1}{l}{MRR} & \multicolumn{1}{l}{RP@1} & \multicolumn{1}{l}{RP@5} & \multicolumn{1}{l}{RP@10} \\ \hline
Decorte et al. \cite{negsample} (best approach) & 0.299                   & N/A                      & 30.82                    & \multicolumn{1}{c|}{38.69}                   & 0.339                   & N/A                      & 31.71                    & 39.19                     \\
Classifiers (no rerank)        & 0.326                   & \textit{\textbf{27.20}}  & \textit{\textbf{37.60}}  & \multicolumn{1}{c|}{\textit{\textbf{46.47}}} & 0.299                   & 27.16                    & 33.41                    & 39.86                     \\
Similarity (no rerank)         & \textit{\textbf{0.355}} & 26.44                    & 35.22                    & \multicolumn{1}{c|}{43.73}                   & \textit{\textbf{0.405}} & \textit{\textbf{32.84}}  & \textit{\textbf{49.67}}  & \textit{\textbf{58.66}}   \\ \hline
GPT3.5 Re-ranking              &                         &                          &                          & \multicolumn{1}{c|}{}                        &                         &                          &                          &                           \\
+Classifier                    & 0.232                   & 18.32                    & 24.10                    & \multicolumn{1}{c|}{27.94}                   & 0.279                   & 21.95                    & 29.30                    & 32.48                     \\
+Similarity                    & 0.369                   & 29.39                    & 34.40                    & \multicolumn{1}{c|}{38.93}                   & 0.413                   & 35.01                    & 43.26                    & 47.15                     \\
+Both                          & 0.372                   & 27.02                    & 32.93                    & \multicolumn{1}{c|}{38.68}                   & 0.369                   & 29.67                    & 37.55                    & 43.24                     \\
+Both + Python                 & \textit{\textbf{0.427}}                   & \textit{\textbf{36.92}}  & \textit{\textbf{43.57}}  & \multicolumn{1}{c|}{\textit{\textbf{51.44}}} & \textbf{\textit{0.488}}                   & \textit{\textbf{40.53}}  & \textit{\textbf{52.50}}  & \textit{\textbf{59.75}}   \\ \hline
GPT4 Re-ranking                &                         &                          &                          & \multicolumn{1}{c|}{}                        &                         &                          &                          &                           \\
+Classifier                    & 0.446                   & 37.16                    & 48.40                    & \multicolumn{1}{c|}{53.44}                   & 0.442                   & 39.10                    & 46.77                    & 51.70                     \\
+Similarity                    & 0.467                   & 36.40                    & 48.35                    & \multicolumn{1}{c|}{54.52}                   & 0.481                   & 40.82                    & 54.15                    & 62.71                     \\
+Both                          & \textbf{0.507}          & \textbf{42.91}           & \bfseries{56.67}           & \multicolumn{1}{c|}{60.09}                   & 0.512                   & 45.67                    & 59.47                    & 64.03                     \\
+Both + Python                 & 0.495                   & 40.70                     & 53.34                    & \multicolumn{1}{c|}{\textbf{61.02}}          & \textbf{0.537}          & \textbf{46.52}           & \textbf{61.50}           & \textbf{68.94}           
\end{tabular}
\caption{Results for the various skills matching approaches. Best results within a category in \textit{\textbf{italicised bold}}, best overall results in \textbf{bold}}
  \label{results}
\end{tablecustom*}

The results of our experiments are presented in in Table \ref{results}. We report the performance of the full pipeline, with both GPT 3.5 and GPT 4 re-ranking, as well as the previous state-of-the-art performance obtained by Decorte et al. in the paper introducing the dataset \cite{negsample}. We also report the results of both our \textbf{Classifier} and \textbf{Similarity} approaches without the re-ranking step, both to showcase the performance obtained via the use of LLM-generated training data and to serve as a baseline for the re-ranking approaches.

We notice that, on their own, both of these no-reranking approaches achieve competitive performance against previous methods, with the similarity approach marginally outperforming the classifier one on the \textbf{House} dataset but performing noticeably worse on the \textbf{Tech} one. These results are encouraging, as they require no real-world training data and are extremely fast at inference-time, requiring only simple computations. Their \textbf{RP@k} scales particularly well with higher \textit{k} values, highlighting their ability to propose a number of correct labels but not ranking them optimally.

\textsc{GPT-4} reranking results in considerable improvements over all non-reranked methods, strongly outperforming all other methods in all approaches and the best-performing variant reaching an RP@10 of \textbf{61.02} on the \textbf{House} dataset and \textbf{68.94} on the \textbf{68.94}, a respective improvement of \textbf{22.33} and \textbf{29.75} percentage points over the previous best approach and \textbf{14.55} and \textbf{10.28} over our best non-reranked methods. We notice that in all cases, the performance obtained by combining potential skills generated by both the classifier and the similarity approaches is noticeably stronger than when using only one method of generating candidates. However, when using a single method of generating potential candidates, we notice that the similarity-based approach tends to outperforms the classifier-based approach on both datasets, especially on the \textbf{House} dataset.

The performance of \textsc{GPT-3.5} re-ranking is more mixed. With natural language prompting, its performance is an overall downgrade over the non-reranked approaches. Unlike \textsc{GPT-4}, we also notice that combining both methods of potential skill generation does not systematically improve performance, especially on the \textbf{Tech} dataset where using only similarity-based entries resulted in overall stronger results. When using only the classifiers-based candidates, we notice that the GPT-3.5 ranking actually decreases performance. One of the noticeable reasons for this weaker performance is GPT-3.5's seemingly weaker ability to follow guidelines: despite our experiments in modifying the model prompt, it would frequently "hallucinate" skills whose wording was directly inspired from the target span, and ranking them higher than the skills provided.

For both GPT variants, we notice strong performance with the \textbf{Python} prompt variation, where we explicitly request that the model output is a Python function returning the ranked list of skills. In the case of GPT-4, the Python variant significantly outperforms natural language prompting on the \textbf{Tech} dataset, but performs slightly worse on \textbf{House} for all metrics but RP@10. For GPT-3.5, however, Python prompting nearly entirely eliminates the problem of hallucinating skills, and greatly improves the performance across all metrics on both datasets. This appears to suggest that framing the problem as a programming problem, which are frequently used to train LLMs, helps ground reasoning and improve performance in re-ranking tasks in a way natural language prompt engineering cannot, although more experiments are needed to confirm this.

Overall, the use of LLM-generated training data out-performs the state-of-the-art distant supervision approaches, and that zero-shot LLM re-ranking further increases performance, considerably outperforming all previous approaches.

\section{Limitations and Future Work}

While our work shows very strong potential for LLMs in both generating training data and improving inference-time predictions for skills matching, we believe that there are three key limitations to our work that should be explored in future work.

\textbf{Broader Scope} We focus on a small, focused dataset which has previously been explored in the literature. We believe that our approach is likely to generalise well to both other taxonomies and different datasets relying on ESCO. We believe future work should explore building upon this method to explore more data sources and evaluation approaches.

\textbf{Representation Types} Our study explores only the use of e5 \cite{e5} embeddings, due to their very strong out-of-the-box performance. However, these embeddings are general domain representations and are only one approach among many. We believe future work exploring different approaches to representation could yield better results and valuable insight. Notably, further exploring techniques common within the field of information retrieval, utilising powerful cross-encoders such as ColBERT, and combining deep-learning based forms of representations with simpler but powerful approaches such as tf-idf capturing different kinds of information could prove very valuable.

\textbf{LLMs Used} This work uses the GPT family of model, and more specifically, GPT-4. These models are gated behind APIs and their weights are not publicly available. While they perform well, future work should explore the applicability of open-source LLMs, such as Falcon \cite{falcon}, as well as look for more efficient approaches. Additionally, we intend to explore if using a more diverse set of generative models, trained on different datasets, could improve our synthetic training data generation by generating more semantically varied examples.

\textbf{Domain-Specific Models} Our approach focuses on the use of general domain model, with no further training to adapt them to the language used within job postings specifically. While we believe that this kind of information is present within the training corpora of the large language models we use, we believe that better targeted models could facilitate the development of more efficient approaches as mentioned above. Notably, models such as JobBERT \cite{jobbert} and ESCOXLM-R \cite{escoxlmr} have shown the potential of domain-specific fine-tuning on existing tasks. Meanwhile, the LLM literature highlights how considerably smaller language models, with an order of magnitude fewer parameters than GPT-3, can reach competitive performance through fine-tuning on small but very high quality datasets \cite{orca, vicuna}.

\textbf{Potential Skills Generation} Our experiments indicate that varying the number of potential skills given to GPT-4 does not have a major impact, if any, on its ranking performance. However, we did not extensively experiment with different ways of generating the potential skills list, and the impact that prompt modifications, such as different ordering or indicating the source or probability given to a label by the initial classifier would have. Additionally, we conducted only very moderate experimentation in optimising the hyper-parameters of our classifier-based candidate generation (as described in Appendix \ref{app:B}) or with alternate ways of computing similarity, such as using SVM-based retrievers \cite{svmretrievers}. We plan to explore these optimisations in future work.

\textbf{Impact on Recommender Systems} While out of the scope of this study, one of the key goals of better job/skill matching is facilitating the use of recommender systems to highlight good matches between jobseekers and job postings, thus contributing to alleviating the job/skill mismatch \cite{skillmismatch}. Our early results in using the output of the pipeline introduced in this paper have shown promising results, and we intend to further explore the best use of this skills extraction pipeline within end-to-end job recommender systems in future work.

\section{Conclusion}

In this work, we have proposed a novel end-to-end zero-shot pipeline for skills matching against the ESCO taxonomy using Large Language Models (LLMs). We have shown that LLMs can generate high-quality synthetic training data to improve candidate generation, outperforming existing approaches without needing any non-synthetic training data. We have also demonstrated that state-of-the-art LLMs can act as strong zero-shot re-rankers as the final step of the skill matching pipeline, resulting in another large performance improvement.

Our experiments also highlight that framing the re-ranking task as a mock Python programming problem results in significant performance gains, especially for less capable models. We believe that this framing helps the models better follow the task instructions in re-ranking contexts, especially when working with less powerful models.

Overall, our work highlights the strong potential for Large Language Models for the low-resource context of working with the ESCO taxonomy, through leveraging the limited information present in the taxonomy to guide the generation of targeted synthetic data, as well as through zero-shot application of their capabilities. While our experiments have focused on a single dataset and taxonomy, namely ESCO, we believe that our approach holds potential to support further work in automated understanding of the job market at scale, and we release the prompts we have used in order to support these efforts.

%% Define the bibliography file to be used
\bibliography{sample-ceur}

% %%
% %% If your work has an appendix, this is the place to put it.
\appendix

\section{Prompts} \label{app:A}

\subsection{Training Data Generation}

You are the leading AI Writer at a large, multinational HR agency. You are considered as the world's best expert at expressing required skills and knowledge in a variety of clear ways. You are particularly proficient with the ESCO Occupation and Skills framework. As you are widely lauded for your job posting writing ability, you will assist the user in all job-posting, job requirements and occupational skills related tasks.

You  work in collaboration with ESCO to gather rigid standards for job postings. Given a list of ESCO skills and knowledges, you're asked to provide forty examples that could be found in a job ad and refer to the skill or knowledge component. You may be given a skill family to help you disambiguate if the skill name could refer to multiple things. Ensure that your examples are well written and could be found in real job advertisement.

Write a variety of different sentences and ensure your examples are well diversified. Use a variety of styles. Write examples using both shorter and longer sentences, as well as examples using short paragraphs of a few sentences, where sometimes only one is directly relevant to the skill. You're trying to provide a representative sample of the many, many ways real job postings would evoke a skill.

At least \{FIVE for tech skills, ZERO for language skills, 80\% (THIRTY-TWO)\} of your examples must not contain an explicit reference to the skill and must thus not contain the given skill string.
Extra Information/Alternative Names (you may discard this information if irrelevant): \{ALTERNATE NAMES IN THE ESCO DATABASE\}
Avoid explicitly using the wording of this extra information in your examples.
Skill: \{target\}"""

\subsection{Reranking}

% \begin{quote}
\textit{\textbf{Instructions}}: You are given an extract from a job posting. As an AI job and skills expert, you need to assist in whatever task is requested of you.
I will give you a sentence referring to a skill extracted from a job posting, as well as a list of potential skill labels. You are asked to extract and rank the likely skills from the candidates list into a ranked list of 10.

It is possible that none match, in which case you will say NO\_LABEL. You must either use one from the list or NO\_LABEL.

You may not use any label not provided in the example list. If you use NO\_LABEL, do not assign any other label.

You will rank the top 10 most likely labels from the candidates, and provide an explanation as to why they are picked and ranked where they are.

That means that if two labels are applicable, but one is much broader, you should pick the less broad one slightly above the broader one. For example, a skill related to specific kind of algorithm (e.g. forecasting) should always rank higher than the "algorithms" skill.

Again, you may never use a skill not provided in the potential skills list.

First, acknowledge and quickly summarise the instructions.

\textit{\textbf{Mocked LLM Message:}} I understand the instructions. I will be given a sentence referring to a skill from a job posting and a list of potential skill labels. My task is to extract and rank the top 10 most likely skills from the candidates list, provide an explanation for my choices, and prioritize specific skills over broader ones. If no label matches, I will use NO\_LABEL and not assign any other label. I will only use the provided labels in the list.

\textit{\textbf{Query Message:}} Great, let's go :)!

Potential skills:
\{\{potential\_skills\}\}

Extract: \{\{text\_extract\}\}

Provide your 10 most likely labels.Please provide a quick explanation for your ranking.
% \end{quote}

\subsection{Python Reranking}

\textit{\textbf{Instructions}}: You are given an extract from a job posting. As an AI job and skills expert, you need to assist in whatever task is requested of you.
I will give you a sentence referring to a skill extracted from a job posting, as well as a list of potential skill labels. You are asked to extract and rank the likely skills from the candidates list into a ranked list of 10.

It is possible that none match, in which case you will say NO\_LABEL. You must either use one from the list or NO\_LABEL.

You may not use any label not provided in the example list. If you use NO\_LABEL, do not assign any other label.

You will rank the top 10 most likely labels from the candidates, and provide an explanation as to why they are picked and ranked where they are.

That means that if two labels are applicable, but one is much broader, you should pick the less broad one slightly above the broader one. For example, a skill related to specific kind of algorithm (e.g. forecasting) should always rank higher than the "algorithms" skill.

Again, you may never use a skill not provided in the potential skills list.

\textbf{You will answer not by directly answering, but by writing a python function that will output your ordered list. Your python function can use any heuristic you wish to reach your desired result and accommodate your reasoning.}

First, acknowledge and quickly summarise the instructions.

\textit{\textbf{Mocked LLM Message:}} \textbf{The task is to create a Python function that will rank the top 10 most likely skills from a list of potential skills based on a given sentence from a job posting. The function should only use the provided list of potential skills and may use the NO\_LABEL option if none of the skills match. The ranking should be based on the specificity of the skill and the explanation for the ranking should be provided.}

\textit{\textbf{Query Message:}} Great, let's go :)!

Potential skills:
\{\{potential\_skills\}\}

Extract: \{\{text\_extract\}\}

Provide your 10 most likely labels.Please provide a quick explanation for your ranking. \textbf{You must provide a Python function. The python function must be in a python codeblock. The function must be named `rank\_skills`. The Python Function you create should never explicitely use the extract text itself, it should just return the ranking. The extract text is only provided for your reasoning so you can build the appropriate function. You don't need to include the full list of potential skills in your answer}

\section{Classifier Parameters} \label{app:B}

The individual logistic regression classifiers are implemented using the scikit-learn library \cite{sklearn}. We set the inverse regularisation parameter, \textit{C} to \textbf{0.1}, as we have low confidence in our data being representative of real-world data, set a maximum iteration limit of \textbf{10 000} with a tolerance of \textbf{0.00001}. We also set the class weight to be used by the classifier to the \textit{balanced} setting, meaning that positive examples will be weighed twice as heavily as negative examples by the loss function, as our negative sampling strategy involves two negative examples per positive one.

\end{document}